\pdfoutput=1

\documentclass[11pt]{article}

\usepackage[final]{coling}

\usepackage{times}
\usepackage{latexsym}

\usepackage[T1]{fontenc}

\usepackage[utf8]{inputenc}

\usepackage{microtype}

\usepackage{inconsolata}

\usepackage{graphicx}

\usepackage{times}
\usepackage{latexsym}
\usepackage{amsmath}
\usepackage{caption}
\usepackage{subcaption}
\usepackage{xurl}

\usepackage[T1]{fontenc}

\usepackage[utf8]{inputenc}

\usepackage{microtype}

\usepackage{inconsolata}

\usepackage{graphicx}

\usepackage{todonotes}

\title{Is Peer-Reviewing Worth the Effort?}

\author{Kenneth Church, Raman Chandrasekar, John E. Ortega and Ibrahim Said Ahmad \\
Institute for Experiential AI, Northeastern University \\
\texttt{\{k.church, r.chandrasekar, j.ortega, i.ahmad\}@northeastern.edu}}

\begin{document}
\maketitle
\begin{abstract}
How effective is peer-reviewing in identifying important papers?
We treat this question as a forecasting task.  Can we predict
which papers will be highly cited in the future based on venue
and ``early returns'' (citations soon after publication)?
We show early returns are more predictive than venue.
Finally, we end with constructive suggestions to address scaling challenges: (a) too many submissions and (b) too few qualified reviewers.
\end{abstract}

\section{Introduction}
\subsection{Prioritization as a Forecasting Task}

\begin{figure}[!ht]
\centering
  \includegraphics[width=0.75\columnwidth]{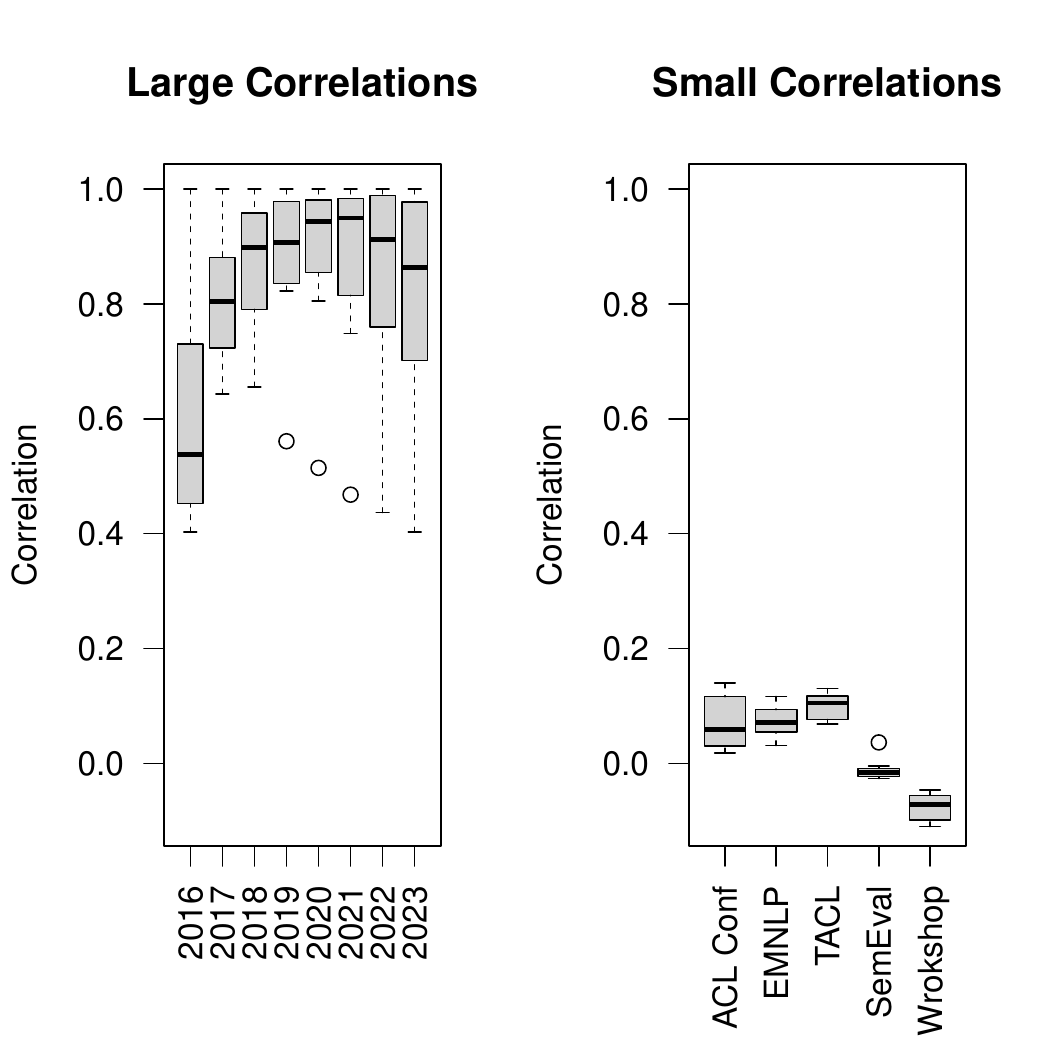}
  \caption{Early Returns (left) $\gg$ Venue (right), based on correlations ($\rho$) from Tables \ref{tab:cor_of_citations}-\ref{tab:cor_of_venues}.  Data is based on Semantic Scholar (S2) \cite{Wade2022TheSS}, where the venue field refers not only to conferences, but also to journals and more.}  
  \label{fig:cor}
\end{figure}

How effective is peer-reviewing in identifying important papers?
Since readers cannot afford to read everything, should they prioritize
papers in top venues, or something else?
Following \citet{davletov2014high,ma2021deeplearning}, we treat this question as a forecasting task.  Can we predict
which papers will be highly cited in the future?  Both venue
and ``early returns'' (citations soon after publication) are statistically significant,
but early citations have larger correlations with future citations as shown in \autoref{fig:cor}.  This figure will be explained in more detail
in \autoref{sec:citation_counts}.  Data for figures and tables
is posted on GitHub.\footnote{\url{https://github.com/kwchurch/is-peer-reviewing-worth-the-effort}}

\citet{abramo2019predicting} also found  early citations
to be more predictive than venue (impact): ``{the role of the impact factor in the combination becomes negligible
after only two years from publication}.''





\subsection{H-Index and Impact}
In some organizations, authors are encouraged to publish in top tier venues, using statistics such as
h-index \cite{hirsch2005index} and impact \cite{Garfield2006TheHA} to rank authors, venues, countries \cite{hyland_2023} and more.  We
use similar summary statistics to show that conditioning on early citations
is more effective than conditioning on venue.
That is, we group papers by venue and by early citations (one year after publication), and summarize citations for the fourth year after publication
with $h$ (h-index) and $\mu$ (impact).  
Results do not depend too much on the details of these definitions of early and future citations because citations are highly correlated over time

When we discuss
Tables~\ref{tab:pubmed_stats}-\ref{tab:stats}, 
$h$ and $\mu$ are better for papers conditioned on early citations
than for papers conditioned on venue.  In particular,
papers in less selective venues (Workshops/ArXiv)
with a few early citations tend to have more citations in the future
than papers in more selective venues.

In addition to $h$ and $\mu$, Tables~\ref{tab:pubmed_stats}-\ref{tab:stats}
 report $N$ (number of papers in each group) and $\sigma$ (standard deviation).  $N$ will be used in discussions of inclusiveness
and $\sigma$ will be used in discussions of robustness.
We will suggest prioritizing papers with early citations is
more effective than prioritizing by venue:

\begin{enumerate}
    \setlength{\itemsep}{0pt}
    \setlength{\parskip}{0pt}
    \setlength{\parsep}{0pt}
    \item More selective: $\rho$ (correlation), $h$, $\mu$ (impact)
    \item More inclusive: $N$ (number of papers)
    
\end{enumerate}
\noindent
These observations are robust, as will be shown
by replications over a number of conditions including papers from different sources (ACL, PubMed, ArXiv) 
and papers with different publication dates.




\section{Related Work}


\subsection{Metrics: H-index and Impact Factor}


There is considerable work on 
metrics of success such as impact factor \cite{Garfield2006TheHA}
and h-index \cite{hirsch2005index}.  Both of these
summary statistics are computed over a group of papers, where papers
are typically grouped by author or by venue, 
depending on whether one is interested in measuring success by author or by venue.
We will group papers in additional ways such as papers with $T$ or more citations
in the first year after publication in order to compare scores
of success by early returns with scores by
other factors such as venue.

Impact factor, $\mu$, is simply the average of citation counts for papers in the group,
and h-index, $h$, is the number of papers in the group with $h$ or more citations.
Many journals report impact factors.  Google Scholar ranks venues by h5, a variant of
h-index, computed over the last five years.
In addition to top venues,\footnote{\url{https://scholar.google.com/citations?view_op=top_venues}}
Google also provides details for many fields such as Computational Linguistics.\footnote{\url{https://scholar.google.com/citations?view_op=top_venues&vq=eng_computationallinguistics}}

\subsection{Numerous Challenges to Reviewing}

The peer-review process, despite being an integral part of academic scholarship, has been a subject of criticism on multiple fronts \cite{tom_slr_2002}:
\begin{quote}
\textit{ the practice of peer review is based on faith in its effects, rather than on facts.}   
\end{quote}
\noindent
In this work, we assume reviews and other assessments of value should be leading indicators of
future citations, following suggestions we have made elsewhere \cite{church-2005-last,church2020emerging}.  While this assumption may be controversial,
it provides an objective path forward.
There are, of course, numerous challenges in reviewing processes;
the first three challenges below are discussed in Sections \ref{sec:purpose}-\ref{sec:cheating}; scale/growth is discussed in \ref{sec:advice}.
\begin{enumerate}
    \setlength{\itemsep}{0pt}
    \setlength{\parskip}{0pt}
    \setlength{\parsep}{0pt}
    \item Poorly defined tasks/incentives  
    \item Validity and Reliability 
        \item Vulnerabilities, Cheating and Ethics 
        \item Scale: Exponential growth
    \item Subjectivity/Biases \cite{lee_bias_2013,huber_2022,smith2023peer} 
    \item Time and Cost \cite{de2009exploring}
\end{enumerate}

\subsubsection{Purpose of peer-reviewing?}
\label{sec:purpose}

What is the purpose of peer-reviewing? The task is not very well-defined.  According to \citet{rogers-augenstein-2020-improve}, ``{reviewers and
area chairs face a poorly defined task 
forcing
apples-to-oranges comparisons}.''
An evaluation of biomedical research publications \cite{chauvin2015most} concluded:
``The most important tasks for peer reviewers were not congruent with the tasks most often requested by
journal editors in their guidelines to reviewers.''

\subsubsection{Validity and Reliability}
\label{sec:validity}

There is considerable discussion of validity and reliability in Experimental Psychology \cite{krippendorff2018content}.  Evaluations of the
reliability of peer-reviewing are worrisome. \citet{Cortes2021InconsistencyIC}
revisited an experiment based on NIPS-2014 (now known as NeurIPS): ``{From the
conference 10\% of the papers were randomly chosen to be reviewed by two independent program committees... results
showed that the decisions between the two committees was better than random, but still surprised the
community by how low it was.}''

The follow up study looked at review scores and future citations.
They failed to find a significant correlation for accepted papers (their figure 6).  For rejected papers that appeared elsewhere, the correlation was not large (their figure 8).

A recent evaluation
of reviews \cite{Goldberg2023PeerRO} found
``many problems that exist
in peer reviews of papers—inconsistencies, biases, miscalibration, subjectivity—also exist in peer reviews of
peer reviews.''






\subsubsection{Vulnerabilities, Cheating and Ethics}
\label{sec:cheating}

There are opportunities for authors, reviewers and other parties to use/abuse chatbots.  A number of funding agencies (NIH\footnote{\url{https://nexus.od.nih.gov/all/2023/06/23/using-ai-in-peer-review-is-a-breach-of-confidentiality/}} and ARC\footnote{\url{https://www.arc.gov.au/sites/default/files/2023-07/Policy\%20on\%20Use\%20of\%20Generative\%20Artificial\%20Intelligence\%20in\%20the\%20ARCs\%20grants\%20programs\%202023.pdf}})
and journals (Science, Lancet, JAMA)
discourage/prohibit reviewers from uploading manuscripts to
AI platforms that cannot guarantee confidentiality \cite{cheng2024generative}.

Even before chatbots, much has been written about ethics and peer-reviewing: \cite{rockwell2006ethics,souder2011ethics,remuzzi2023ethics}.
There have always been many ways to cheat.  Advances in technology create new and better ways to cheat, as well as new and better ways to catch cheating.

In this work, we will use citations, which admittedly can be purchased/gamed\footnote{\url{https://www.science.org/content/article/vendor-offering-citations-purchase-latest-bad-actor-scholarly-publishing}} \cite{Beel2010AcademicSE}.
Spam is obviously a cat-and-mouse game but purchasing citations is unlikely to be
successful for long.  Given the correlations over time,
cheaters would need to purchase citations for many years or else it is too easy
to catch them by looking for
anomalies in citation counts over time. 
Moreover, with h-index, it is too easy to find the small number
of papers that contribute to the score.
There are easier and more effective ways to cheat such as plagiarism
and chat bots.

\subsection{Related Work on Predicting Citations}

This paper questions whether peer-reviewing
is worth the effort.
Prior work is more about improving predictions (\autoref{sec:improving}),
or helping authors
increase their citations (\autoref{sec:advice}).

\subsubsection{Improving Predictions}
\label{sec:improving}

There is a considerable
body of work on 
predicting citations.
Predicting citations can be viewed as a special case
of time series prediction.  There are many use cases, especially in finance: 
\cite{salinas2020deepar}.
Prior work often focuses on methods:
linear regression \cite{Pobiedina2016CitationCP},
negative binomials\footnote{Negative binomials are a natural choice for highly skewed data.
Citations tend to be highly skewed as can be seen from standard deviations ($\sigma$) in many
of the tables in this paper.  If citations were generated by a Poisson process,
then ${\sigma}^2 \approx \mu$, but citations have long tails where $\sigma \gg \mu$ (in most cases).} \cite{onodera2015factors}, clustering \cite{davletov2014high}, nearest neighbors \cite{yan2011citation} and deep networks \cite{abrishami2019predicting,ruan2020predicting}.
There is considerable work on link prediction in the literature on GNNs (graph neural networks) using the ogbl-citation2 task in OGB (Open Graph Benchmark) \cite{Hu2020OpenGB}.
In more recent work, \citet{dewinter2024canchatgpt} aims to ``pave the way for AI-assisted peer review,'' using ChatGPT4 to analyze 2222 abstracts with 60 criteria. Using principal component analysis, three components are identified, of which two -- about Accessibility \& Understandability, and Novelty \& Engagement, are linked to citation counts.  





In addition to methods for predicting citations, there are also
discussions of features:

\begin{enumerate}
    \setlength{\itemsep}{0pt}
    \setlength{\parskip}{0pt}
    \setlength{\parsep}{0pt}
    \item Early Citations: \citet{Wang2013QuantifyingLS,davletov2014high,abramo2019predicting,Bai2019PredictingTC,stegehuis2015predicting,ma2021deeplearning,pengwei2024modeling}
    \item Venue: \citet{yan2011citation,abramo2019predicting}
    \item Properties of authors: author rank, h-index, productivity, etc. \citet{yan2011citation}
    \item Contents of paper: \citet{huang2022finegrained} predict citations 
    based on sections (introduction, background, method, etc.)  of a paper.
    \end{enumerate}

\subsubsection{Advice to Authors}
\label{sec:advice}

There is considerable advice to authors
on how to increase citations.
We have argued elsewhere \cite{church2017word2vec} that
secondary sources are cited more than primary sources;
the most cited papers often help others make progress, e.g., datasets, GitHubs, models on HuggingFace, benchmarks, tools, surveys, textbooks.
By construction, the last word on a topic
is not cited.  
The most cited paper is rarely
the first, last or best; simplicity and accessibility are
preferred over timing and quality.

\citet{tahamtan2016factors} survey the literature
on advice to authors, assigning prior work to 28 factors,
which we have aggregated/condensed down to 8.  Their 28 factors
seem plausible, though it is not possible to discuss all 28 factors in this paper.

\begin{enumerate}
    \setlength{\itemsep}{0pt}
    \setlength{\parskip}{0pt}
    \setlength{\parsep}{0pt}
    \item Intrinsic properties of paper: quality, length, number of references.
    Figures, charts and appendices can increase citations, but challenging equations can decrease citations.
    \item Venue: metrics ($\mu$, h), prestige, language.
    \item Discipline/subject/topic/methodology
    \item Accessibility and visibility of papers: Avoid pay walls \cite{lawrence2001free,eysenbach2006citation}, and promote papers on social media/ArXiv.
    \item Primary Source vs. Secondary Source: Textbooks and survey papers are highly cited, as are tools, benchmarks and datasets.
    \item Demographics of author(s): Number of authors, self-citations,
    country, gender, age, reputation, productivity, affiliation, funding.
    \item Publication date: Since the literature is growing exponentially, doubling every 17 years \cite{bornmann2021growth,Redner2005CitationSF},
    papers published recently tend to have more citations.
    \item Early citations: Citations soon after publication are predictive of future citations, though
    there are exceptions such as ``Sleeping Beauties'' \cite{Raan2004SleepingBI}.
\end{enumerate}

\begin{table}[hb!]
  \centering
  {\small
  \begin{tabular}{ r r r r r r r r r }
\textbf{Venue} & \textbf{Id in S2} &  \rotatebox{90}{\textbf{2016}}&
\rotatebox{90}{\textbf{2017}}&
\rotatebox{90}{\textbf{2018}}&
\rotatebox{90}{\textbf{2019}}&
\rotatebox{90}{\textbf{2020}}&
\rotatebox{90}{\textbf{2021}}\\ \hline
      NAACL & 9724599 & 5 & 7 & 5 & 1 & 3 & 1  \\
       LREC & 12260053 & 0 & 0 & 0 & 1 & 0 & 0  \\
  LREC & 28309452 & 2 & 8 & 4 & 10 & 7 & 7  \\
  EMNLP & 1380793 & 0 & 2 & 16 & 19 & 17 & 19  \\
  COLING & 18649702 & 0 & 1 & 2 & 1 & 3 & 1  \\
  SemEval & 17378758 & 0 & 0 & 0 & 2 & 0 & 0 \\
    \end{tabular}}
  \caption{Citation counts from Semantic Scholar (S2) for a few ACL papers published in 2016.}
  \label{tab:citation_Counts}
\end{table}

 \begin{table}
  \centering
  {\small
  \begin{tabular}{ c | c c c c c  c}
    & \rotatebox{0}{\textbf{2016}}&
\rotatebox{0}{\textbf{2017}}&
\rotatebox{0}{\textbf{2018}}&
\rotatebox{0}{\textbf{2019}}&
\rotatebox{0}{\textbf{2020}}&
\rotatebox{0}{\textbf{2021}} \\ 
  & \multicolumn{6}{c}{\textbf{3710 ACL Papers Pub. in 2016}} \\
\hline
2016& 1.00& 0.80& 0.66& 0.56& 0.51& 0.47 \\
2017& 0.80& 1.00& 0.92& 0.85& 0.81& 0.75\\
2018& 0.66& 0.92 &1.00 &0.98 &0.94 &0.88  \\
2019 &0.56 &0.85 &0.98& 1.00& 0.98& 0.93\\
2020& 0.51& 0.81& 0.94& 0.98& 1.00& 0.98 \\
2021& 0.47& 0.75& 0.88& 0.93& 0.98& 1.00 \\
2022& 0.44& 0.70& 0.82& 0.88& 0.95& 0.99 \\
2023& 0.40& 0.64& 0.76& 0.82& 0.90& 0.97 \\
  & \multicolumn{6}{c}{\textbf{1,026,798 PubMed Papers Pub. in 2016}} \\ \hline
2016 & 1.00 & 0.77 & 0.64 & 0.55 & 0.50 & 0.45 \\
2017 & 0.77 & 1.00 & 0.90 & 0.82 & 0.75 & 0.68  \\
2018 & 0.64 & 0.90 & 1.00 & 0.94 & 0.89 & 0.83  \\
2019 & 0.55 & 0.82 & 0.94 & 1.00 & 0.94 & 0.90  \\
2020 & 0.50 & 0.75 & 0.89 & 0.94 & 1.00 & 0.95  \\
2021 & 0.45 & 0.68 & 0.83 & 0.90 & 0.95 & 1.00  \\
2022 & 0.40 & 0.61 & 0.76 & 0.84 & 0.91 & 0.96 \\
2023 & 0.35 & 0.54 & 0.69 & 0.78 & 0.86 & 0.93 \\ \hline
  \end{tabular}}
  \caption{Citation counts (from Semantic Scholar) are highly correlated from one year to the next.}
  \label{tab:cor_of_citations}
\end{table}

\section{Predictions Based on Citations}
\label{sec:citation_counts}

As suggested above, we will use a prediction task
to show that early returns are more effective than venue.
\autoref{fig:cor} is based on citation counts from Semantic Scholar (S2) \cite{Wade2022TheSS}.
For papers in ACL Anthology, PubMed and ArXiv, published between 2016 and 2019,
we extracted citations by year, as illustrated
in \autoref{tab:citation_Counts}.
There are slightly more than a million papers per year in PubMed,
100k/year in ArXiv and 3k/year in ACL.
The next 3 subsections use these citations to:

\begin{enumerate}
    \setlength{\itemsep}{0pt}
    \setlength{\parskip}{0pt}
    \setlength{\parsep}{0pt}
    \item Compute correlations ($\rho$) over time and venue
    \item Compute h-index ($h$) and impact ($\mu$) for papers grouped by early citations and venue 
    \item Forecast citations with regression 
\end{enumerate}
\noindent
All 3 subsections demonstrate that early citations are more predictive of future citations than venue.

\subsection{Correlations}
\label{sec:cor}

The top of \autoref{tab:cor_of_citations} focuses on 3710 ACL papers published in 2016.  The correlation ($\rho$) of 0.80 between 2016 and 2017 compares the citation counts for these 3710 papers in 2016 and 2017.  The bottom of \autoref{tab:cor_of_citations} is similar
except for the source of papers is now 1,026,798 PubMed papers.  Both the top and bottom of \autoref{tab:cor_of_citations}
start with papers published in 2016.
The correlation of 0.80 above between 2016 and 2017 drops slightly to 0.77
for PubMed papers.

\autoref{tab:cor_of_venues} is like \autoref{tab:cor_of_citations}, but for venues.
Venue is a binary indicator variable containing 1 if the paper appears in that venue and 0 otherwise.  
\autoref{fig:cor}
is based on correlations for ACL papers published in 2016.
\autoref{fig:cor} (left) is based on \autoref{tab:cor_of_citations} (top), and \autoref{fig:cor} (right) is based on \autoref{tab:cor_of_venues} (top).

In addition to the main point, there are a number of interesting (though smaller) effects:
\begin{enumerate}
    \setlength{\itemsep}{0pt}
    \setlength{\parskip}{0pt}
    \setlength{\parsep}{0pt}
    \item \textbf{Main point}: Correlations for early returns are much larger than correlations for venue.
    \item \textbf{Prestige}: Top venues (the ACL main conference, EMNLP and TACL)
    have larger correlations with future citations than workshops.
    \item \textbf{Forecasting horizon}: 
    Because short-term forecasting is easier than long-term forecasting,
 correlations closer to the main diagonal of \autoref{tab:cor_of_citations}
    are relatively large.
    \item \textbf{Quantization}: Correlations for 2016 are relatively small because dates are quantized to years.  There are two dates: year of publication and year of citation.  Citation counts for the year of publication are relatively
    small because that is a partial year.
    \item \textbf{Latency}: It takes time for papers to accumulate citations, and therefore, correlations improve for several years after publication.
\end{enumerate}

\noindent
These observations are robust.  Tables~\ref{tab:cor_of_citations}-\ref{tab:cor_of_venues} replicate the correlations for two types of sources
of papers.  The tables below replicate similar observations over two publication dates, using $h$ and $\mu$ instead of $\rho$.

\begin{table}
  \centering
  {\small
  \begin{tabular}{r | c c c c  }

  \multicolumn{5}{c}{\textbf{3710 Papers in ACL Anthology (2016)}} \\
    & \rotatebox{0}{\textbf{2016}}&
\rotatebox{0}{\textbf{2017}}&
\rotatebox{0}{\textbf{2018}}&
\rotatebox{0}{\textbf{2019}} \\ \hline
ACL Conf&  0.140 &  0.136 &  0.096 &  0.068\\
EMNLP&  0.031 &  0.116 &  0.103 &  0.084 \\
TACL &  0.069 &  0.111 &  0.130 &  0.120 \\
SemEval &  0.036 & -0.005 & -0.026 & -0.024  \\
Workshops&  -0.110 & -0.104 & -0.094 & -0.077\\
  \multicolumn{5}{c}{\textbf{1,121,081 Papers in PubMed, ArXiv or ACL (2016)}} \\ \hline
ACL&           0.0086 & 0.0109 & 0.016 & 0.015   \\
ArXiv   &        0.0255 & 0.0088 & 0.024 & 0.021 \\
PubMed   &     -0.0212 & -0.0012 & -0.014 & -0.013\\ \hline
  \end{tabular}}
  \caption{Correlations with venue are smaller.}
  \label{tab:cor_of_venues}
\end{table}


\begin{table*}
{\small
  \centering
  \begin{tabular}{ p{3.5cm} | r r r r r | r r r r r}
   & \multicolumn{5}{c}{\textbf{Published in 2016}} & \multicolumn{5}{c}{\textbf{Published in 2017}} \\ 
    \textbf{Group}    &    \textbf{h} & \textbf{median} & \textbf{$\mu$} &   \textbf{$\sigma$} &       \textbf{N} &   \textbf{h} & \textbf{median} & \textbf{$\mu$} &   \textbf{$\sigma$} &       \textbf{N} \\ \hline

0 citations & 47 & 1 & 1.4 & 2.3 & 265,090 & 33 & 1 & 1.3 & 2.1 & 268,696 \\
1+ citations& 292 & 3 & 6.8 & 19.8 & 761,729 & 298 & 4 & 7.3 & 19.1 & 808,772 \\
2+ citations& 291 & 5 & 8.5 & 22.8 & 557,824 & 298 & 5 & 9.1 & 22.0 & 591,657 \\
3+ citations& 291 & 6 & 10.3 & 26.1 & 414,641 & 298 & 7 & 11.0 & 25.2 & 438,445 \\
10+ citations& 289 & 17 & 27.1 & 54.2 & 81,125 & 297 & 19 & 28.8 & 52.1 & 84,876 \\
20+ citations& 288 & 37 & 56.4 & 97.3 & 21,329 & 295 & 41 & 59.5 & 92.8 & 22,119 \\

 \hline

{\small Science }               &                                                         113 & 8 & 30.3 & 59.9 & 1732 & 97 & 3 & 22.1 & 48.9 & 1829 \\
{\small Nature }                 &                                                        112 & 5 & 28.0 & 74.9 & 2094 & 109 & 4 & 27.1 & 70.7 & 2020 \\
{\small Nature Communications}    &                                                        88 & 12 & 18.7 & 25.9 & 3702 & 85 & 11 & 17.8 & 32.0 & 4505 \\
{\small J. Amer. Chemical Soc.}                                        & 68 & 10 & 15.9 & 22.1 & 2414 & 68 & 8 & 13.8 & 20.0 & 2679 \\
{\small Proc of the Nat. Academy of Sciences of the USA}  & 73 & 9 & 14.6 & 25.0 & 3629 & 68 & 8 & 12.8 & 19.4 & 3846 \\
{\small bioRxiv} & 55 & 6 & 14.1 & 41.9 & 1396 & 54 & 5 & 10.4 & 29.6 & 3121 \\
{\small Angewandte Chemie}                                                                & 70 & 7 & 13.5 & 21.3 & 2776 & 62 & 6 & 10.7 & 15.9 & 2795 \\
{\small Bioresource Technology}                                                           & 42 & 7 & 10.8 & 15.0 & 1612 & 40 & 7 & 10.0 & 11.9 & 1644 \\
{\small ACS Applied Materials...  }                                           & 49 & 7 & 10.0 & 10.7 & 4074 & 50 & 6 & 9.5 & 10.3 & 4933 \\
{\small Nutrients    }                                                                    & 34 & 5 & 9.4 & 16.6 & 825 & 38 & 6 & 8.8 & 14.9 & 1356 \\
{\small Frontiers in Plant Science }                                                      & 44 & 6 & 9.0 & 15.7 & 2045 & 41 & 6 & 8.7 & 14.0 & 2257 \\
{\small Physical Review Letters}                                                          & 48 & 5 & 8.8 & 13.4 & 2501 & 42 & 4 & 6.9 & 11.7 & 2674 \\
{\small Frontiers in Immunology}                                                          & 29 & 5 & 8.6 & 11.4 & 671 & 39 & 5 & 8.0 & 10.3 & 1901 \\
{\small Sci. of the Total Environ.}                                                & 42 & 6 & 8.6 & 11.3 & 2508 & 46 & 5 & 8.9 & 15.2 & 2853 \\
{\small Frontiers in Microbiology}                                                        & 43 & 5 & 8.6 & 14.9 & 2175 & 47 & 5 & 8.2 & 14.2 & 2621 \\
{\small Food Chemistry }                                                                  & 36 & 6 & 8.5 & 9.4 & 1930 & 31 & 6 & 8.2 & 7.8 & 1798 \\
{\small Chemosphere}                                                                      & 35 & 5 & 7.7 & 12.1 & 1653 & 35 & 4 & 7.2 & 11.8 & 1916 \\
{\small Nanoscale}                                                                        & 37 & 4 & 7.2 & 9.3 & 2166 & 31 & 4 & 5.9 & 7.8 & 2147 \\
{\small I.J. Bio. Macromolecules}                               & 28 & 5 & 7.2 & 8.6 & 1159 & 32 & 5 & 7.4 & 9.8 & 1616 \\
{\small I.J. of Molecular Sciences}                                      & 37 & 4 & 7.1 & 11.4 & 2068 & 47 & 4 & 7.9 & 13.3 & 2746 \\
{\small Scientific Reports}                                                               & 67 & 4 & 6.8 & 14.5 & 20,860 & 64 & 4 & 6.3 & 22.2 & 25,006 \\
{\small Analytical Chemistry}                                                             & 29 & 5 & 6.8 & 7.5 & 1648 & 30 & 4 & 6.3 & 7.3 & 1817 \\
{\small BMJ Open}                                                                         & 33 & 3 & 5.9 & 12.0 & 2016 & 31 & 3 & 4.8 & 8.1 & 2554 \\
{\small Molecules}                                                                        & 34 & 3 & 5.8 & 12.2 & 1745 & 35 & 3 & 5.4 & 8.5 & 2247 \\
{\small Frontiers in Psychology}                                                          & 38 & 3 & 5.7 & 9.9 & 2074 & 32 & 3 & 5.5 & 12.2 & 2252 \\
{\small OncoTarget}                                                                       & 42 & 4 & 5.5 & 6.8 & 7454 & 37 & 3 & 4.0 & 6.1 & 9282 \\
{\small Materials}                                                                        & 24 & 3 & 5.3 & 7.8 & 1024 & 27 & 3 & 4.9 & 8.5 & 1473 \\
{\small J. of Biological Chemistry }                                                 & 26 & 4 & 5.2 & 6.5 & 2135 & 27 & 3 & 5.0 & 6.2 & 1852 \\
{\small Chemical Comm.     }                                                    & 33 & 3 & 5.2 & 7.2 & 3046 & 25 & 3 & 3.9 & 4.7 & 2689 \\
{\small Italian Nat. C. on Sensors }                                          & 34 & 3 & 5.2 & 12.9 & 2220 & 39 & 3 & 5.3 & 10.8 & 2962 \\
{\small British medical journal            }                                              & 41 & 0 & 5.2 & 37.2 & 1837 & 39 & 0 & 4.7 & 29.6 & 1670 \\
{\small I.J. Env. Res. and Pub...  }             & 24 & 3 & 5.2 & 7.1 & 1118 & 31 & 4 & 6.2 & 11.1 & 1575 \\
{\small Organic Letters                        }                                          & 18 & 4 & 5.1 & 4.1 & 1646 & 16 & 3 & 4.0 & 3.4 & 1699 \\

{\small PLoS ONE  }                                                                       & 59 & 3 & 5.1 & 8.8 & 22,512 & 55 & 3 & 4.8 & 8.6 & 20,617 \\
{\small Environ. science and...  }                     & 30 & 3 & 4.9 & 6.6 & 2430 & 32 & 3 & 4.9 & 7.5 & 2527 \\
{\small Chemistry  }                                                                      & 26 & 3 & 4.8 & 5.8 & 2271 & 23 & 3 & 3.8 & 4.5 & 2306 \\
{\small BioMed Research Inter.}                                                    & 25 & 3 & 4.5 & 8.8 & 1790 & 27 & 2 & 4.4 & 7.2 & 2005 \\
{\small Medicine      }                                                                   & 27 & 2 & 4.1 & 14.2 & 3275 & 22 & 2 & 2.8 & 8.5 & 3526 \\
{\small Optics Express     }                                                              & 26 & 2 & 3.9 & 5.3 & 2871 & 23 & 2 & 3.4 & 5.0 & 2739 \\
{\small Physical Chem... - PCCP    }                                  & 25 & 2 & 3.6 & 5.4 & 3584 & 22 & 2 & 2.9 & 5.6 & 3258 \\
{\small RSC Advances            }                                                         & 8 & 3 & 3.5 & 3.1 & 78 & 8 & 3 & 4.2 & 5.0 & 60 \\
{\small Biochemical... - BBRC    }                   & 19 & 2 & 3.5 & 5.3 & 1744 & 22 & 2 & 3.6 & 6.6 & 2056 \\
{\small J. of Chemical Physics   }                                                   & 23 & 2 & 3.4 & 7.8 & 2087 & 19 & 2 & 2.8 & 6.8 & 1944 \\
{\small Dalton Transactions       }                                                       & 22 & 2 & 3.4 & 4.5 & 2085 & 18 & 2 & 2.9 & 3.8 & 1791 \\
{\small World Neurosurgery         }                                                      & 15 & 2 & 2.7 & 3.7 & 1300 & 18 & 2 & 2.6 & 3.6 & 1999 \\
{\small Oncology Letters       }                                                          & 15 & 2 & 2.7 & 3.5 & 1517 & 16 & 2 & 2.7 & 3.5 & 2207 \\
{\small Physical Review E          }                                                      & 18 & 1 & 2.5 & 3.5 & 2284 & 18 & 1 & 2.1 & 3.7 & 2172 \\
{\small M. in molecular biology    }                                                 & 18 & 1 & 1.8 & 4.9 & 2888 & 24 & 1 & 2.0 & 5.5 & 3612 \\
{\small BMJ Case Reports      }                                                           & 9 & 1 & 1.1 & 1.7 & 1401 & 7 & 1 & 1.0 & 1.3 & 1689 \\
{\small Zootaxa        }                                                                  & 11 & 0 & 1.1 & 2.2 & 1967 & 9 & 0 & 1.0 & 3.0 & 1224 \\ \hline
{\small \textbf{All other venues }    }                                                                       & 274 & 2 & 5.1 & 17.1 & 878,782 & 273 & 2 & 4.9 & 18.8 & 913,401 \\
 \hline
  \end{tabular}}
  \caption{Deep dive into PubMed papers.  A few early citations compare favorably to most venues.  Early citations are based on first year after publication, and summary statistics are based on fourth year after publication.}
  \label{tab:pubmed_stats}
\end{table*}

\begin{table*}
{\small
  \centering
  \begin{tabular}{ r| r r r r r | r r r r r}
   & \multicolumn{5}{c}{\textbf{Published in 2016}} & \multicolumn{5}{c}{\textbf{Published in 2017}} \\ 
    \textbf{Group}    &    \textbf{h} & {\rotatebox{0}{\small{\textbf{median}}}} & \textbf{$\mu$} &   \textbf{$\sigma$} &       \textbf{N} &   \textbf{h} & {\rotatebox{0}{\small{\textbf{median}}}} & \textbf{$\mu$} &   \textbf{$\sigma$} &       \textbf{N} \\ \hline
  \multicolumn{11}{l}{\textbf{PubMed, ArXiv and ACL Anthology}}\\
    0 citations & 48 & 1 & 1.3 & 2.3 & 292,566 & 35 & 1 & 1.3 & 2.1 & 295,467 \\
1+ citations & 345 & 3 & 6.9 & 24.7 & 828,515 & 339 & 4 & 7.4 & 24.0 & 883,685 \\
2+ citations & 345 & 5 & 8.7 & 28.6 & 604,536 & 338 & 5 & 9.2 & 27.8 & 645,904 \\
3+ citations & 345 & 6 & 10.6 & 32.9 & 448,541 & 338 & 6 & 11.3 & 32.0 & 479,097 \\
10+ citations & 343 & 17 & 28.6 & 70.1 & 88,490 & 337 & 19 & 30.0 & 67.5 & 95,105 \\
20+ citations & 341 & 37 & 61.4 & 127.8 & 23,593 & 336 & 41 & 62.9 & 122.1 & 25,613 \\  \hline
ACL Anthology &     73 &      2 &  9.9 & 59.0 &    3710 &  65  &     2 &  8.5 & 29.1 &    3030 \\
ArXiv &  236 &      2 &  6.4  & 47.0 &   101,176 & 234 &      1 &  6.1 & 58.4 &  110,184 \\
PubMed & 292 &      2 &  5.4 & 17.2 & 1,026,798 & 293 &      2 &  5.2 & 18.6 & 107,7437 \\ \hline
\multicolumn{11}{l}{\textbf{Deep Dive into ACL Anthology}} \\
     0 citations & 9 & 0 & 1.0 & 1.8 & 953 & 7 & 0 & 0.9 & 1.6 & 589 \\
1+ citations & 73 & 3 & 13.0 & 68.2 & 2757 & 65 & 3 & 10.3 & 32.2 & 2441 \\
2+ citations & 73 & 4 & 17.1 & 79.2 & 2025 & 65 & 4 & 12.9 & 36.0 & 1902 \\
3+ citations & 73 & 6 & 21.6 & 90.0 & 1550 & 65 & 5 & 15.7 & 39.8 & 1518 \\
10+ citations & 73 & 23 & 57.0 & 155.6 & 481 & 65 & 18 & 35.3 & 61.3 & 545 \\
20+ citations & 71 & 54 & 114.9 & 235.6 & 190 & 63 & 34 & 62.2 & 86.9 & 223 \\
 \hline

ACL Main Conf.& 42 & 5 & 18.7 & 45.2 & 377 & 41 & 7 & 20.3 & 50.7 & 353 \\
EMNLP & 41 & 6 & 25.8 & 78.9 & 269 & 36 & 6 & 17.9 & 36.5 & 339\\
TACL & 17 & 11 & 70.5 & 280.3 & 45 & 12 & 8 & 15.4 & 23.3 & 41 \\
SemEval & 15 & 1 & 4.7 & 16.3 & 230 & 14 & 1 & 5.5 & 23.5 & 208\\ 
Workshops & 24 & 1 & 3.8 & 10.1 & 1111 & 30 & 1 & 4.2 & 12.5 & 1191\\
\hline
  \end{tabular}}
  \caption{Similar to \autoref{tab:pubmed_stats}, but for papers from different sources.  Note ArXiv is better than PubMed in terms of $\mu$.  
  }


  \label{tab:stats}
\end{table*}

\subsection{H-index and Impact}
\label{sec:h_and_mu}

How can we identify papers that will be highly cited in the future?  The previous
section used correlations ($\rho$).  This section will use h-index ($h$) and impact factors ($\mu$).  Tables~\ref{tab:pubmed_stats}-\ref{tab:stats} group papers based
on citations a year after publication, and report summary statistics of citations
in the fourth year after publication.  \autoref{tab:pubmed_stats} does this
The main observation is: conditioning on papers with early citations compares
favorably to conditioning by venue.

\begin{enumerate}
    \setlength{\itemsep}{0pt}
    \setlength{\parskip}{0pt}
    \setlength{\parsep}{0pt}
    \item
    \textbf{Exclusivity}: Papers with 20+ citations in the first year after
publication are better than all 50 venues in \autoref{tab:pubmed_stats}
in terms of $h$ and $\mu$.  
\item \textbf{Inclusivity}: There are more papers ($N$) with 20+ early citations than in most venues.
\item \textbf{Robustness}: We obtain similar results under a number of conditions including different publication years and different sources of papers.
\end{enumerate}

\begin{figure}[hb!]
\centering
  \includegraphics[width=\columnwidth]{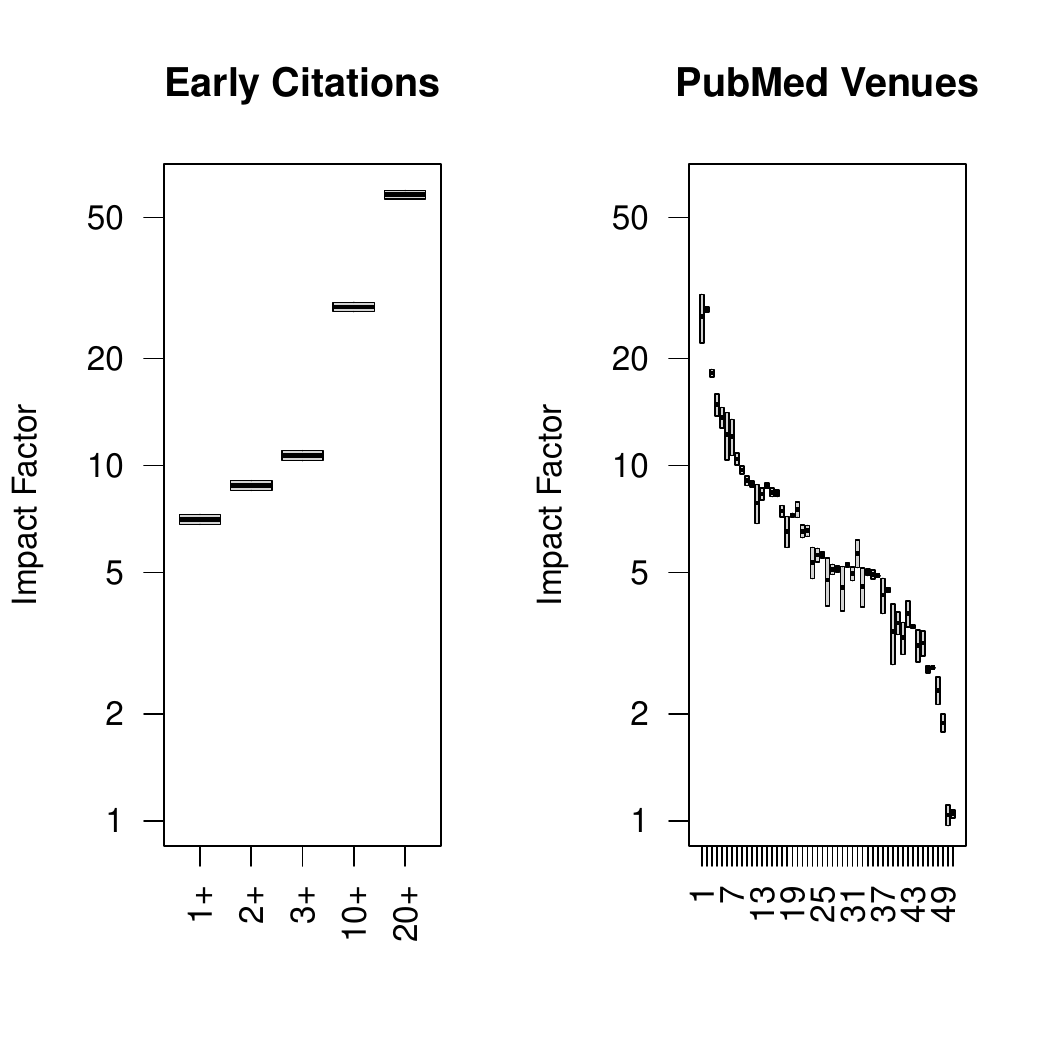}
  \caption{Impact factor ($\mu$) from \autoref{tab:pubmed_stats}.
  Simple rule of thumb: for most venues, reviewers
  are no better than 1+ early citations in terms of $\mu$;
  for all venues, reviewers are no better than 20+ early citations.
  }
  \label{fig:pubmed_stats}
\end{figure}

\autoref{fig:pubmed_stats} plots impact factors ($\mu$) from \autoref{tab:pubmed_stats}, comparing early citations (left)
with 50 PubMed venues (right).  The figure shows that papers with 20+ early citations
have a larger $\mu$ than all 50 venues.  If we select on 1+ early citations, then $\mu \approx 6.8$ is better than 60\% of venues in \autoref{tab:pubmed_stats}.

Note that $h$ does not change much with thresholds on early citations.  That is,
$h$ for 1+ citations is similar to $h$ for 2+ citations because
$h$ is dominated by a few highly cited papers.  As mentioned above, there are a few
``sleeping beauty'' papers that suddenly become highly cited after a few years,
but that is unusual.  The row for 0 early citations shows that papers with no early citations will have a few citations later on ($\mu \approx 1.4 \pm 2.3$).
However, it is more common for papers that will be important
to start
off with more citations early on.

\autoref{tab:stats} is similar to \autoref{tab:pubmed_stats} but for
papers from different sources.
In \autoref{tab:stats}, the row for 3+ early citations is (usually) better than venues in terms of $h$ and $\mu$ with an exception for TACL in 2016,
because of a single highly cited outlier: \cite{bojanowski-etal-2017-enriching}.
  It is risky to average over small samples of highly skewed numbers, as evidence by the large $\sigma$ (standard deviations).  Note that $h$ is more stable than $\mu$ over 2016 and 2017.  
  
  Many ACL venues are highly selective in terms of $\mu$, but ACL could
  improve inclusiveness ($N$) as well as exclusiveness ($\mu$) by publishing more papers from preprint archives such as ArXiv with impressive early citations.
 We will discuss this suggestion in more detail in \autoref{sec:DDI}.



\begin{figure*}
\begin{subfigure}[b]{0.48\textwidth}
  \includegraphics[width=\textwidth]{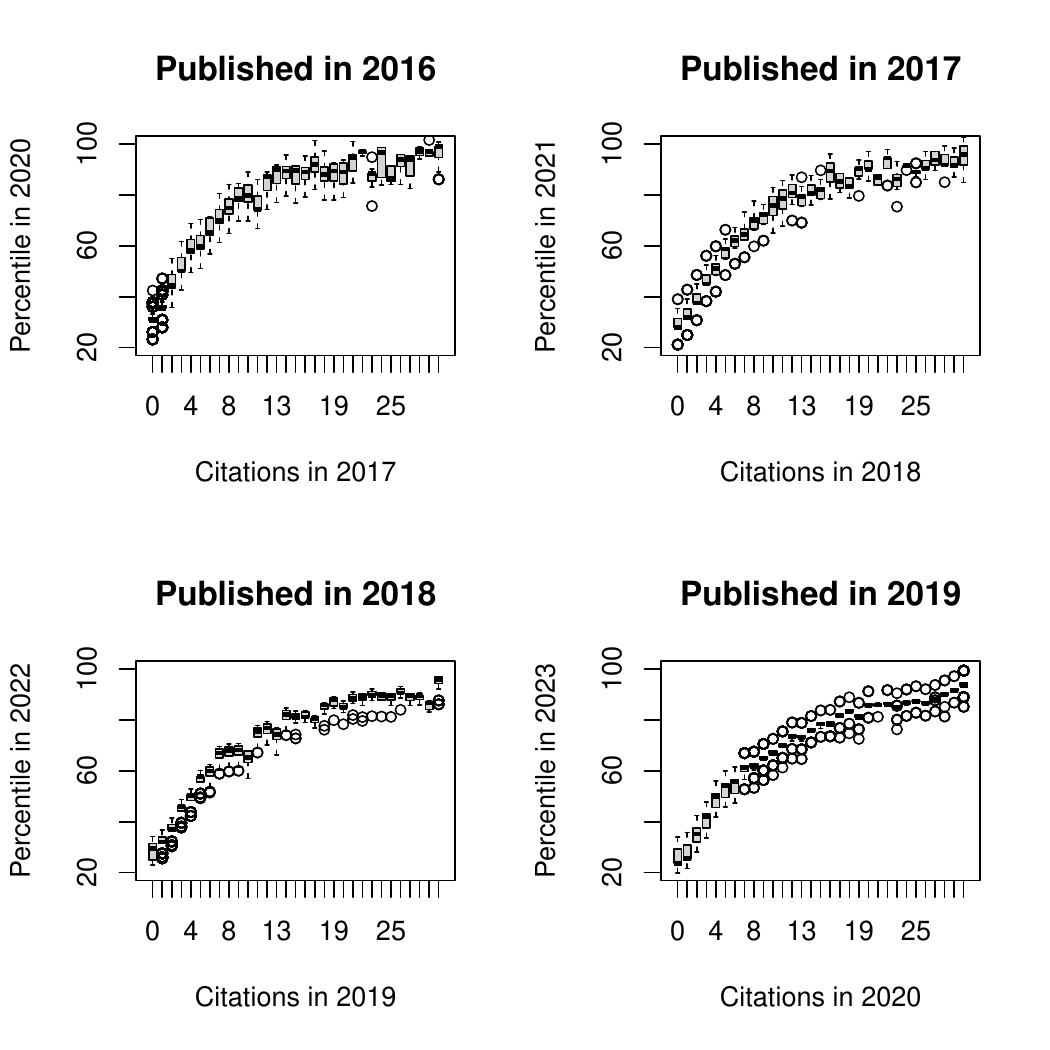}
  \caption{Predictions based on early citations.}
  \label{fig:ACL_citations}
\end{subfigure}
~
\begin{subfigure}[b]{0.48\textwidth}
  \includegraphics[width=\textwidth]{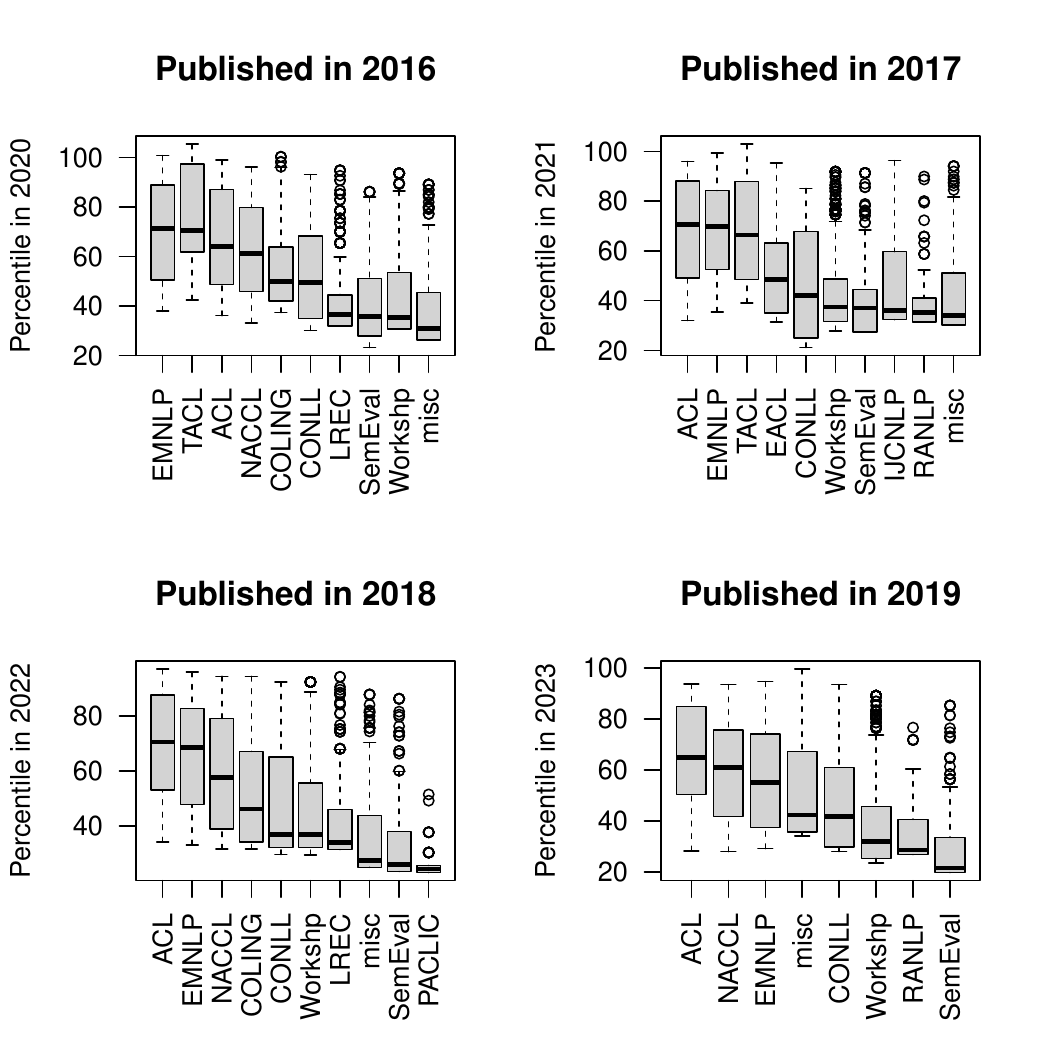}
  \caption{Predictions based on venue.}
  \label{fig:ACL_venue}
\end{subfigure}
  \caption{Boxplots of predictions from regression model for ACL papers.  The bars are so narrow that they are hard to see on the left because early returns are more predictive than venue.}
  \label{fig:ACL}
\end{figure*}

\subsection{Forecasting with Regression}
\label{sec:regression}

We will use the regression model in \autoref{eqn:model} to compare early returns and venue.
\begin{equation}
\label{eqn:model}
\begin{split}
    per&centile_{year+4} \sim venue + \\
    & factor(pmin(T, citations_{year+1}))
    \end{split}
\end{equation}
\noindent
This model predicts the percentile of the paper in the fourth year
based on the venue and early citations.   Early citations
are treated as a factor variable; thus, the model produces
a coefficient for each count between 1 and $T$, as illustrated
in \autoref{tab:coef}.

\begin{table}
{\small
  \centering
  \begin{tabular}{ r | r r | r r }
       & \multicolumn{2}{c}{\textbf{Large Set}}  & \multicolumn{2}{c}{\textbf{Small Set}} \\ 
  Coefficient & 2016 & 2017 & 2016 & 2017 \\ \hline
Intercept & 15.7 & 15.3 & 36.7 & 32.8  \\
ACL Anthology & 4.3 & 1.5  \\
ArXiv & 5.4 & 3.4 \\
PubMed & 10.7 & 11.3 \\
TACL && & 6.8 & 6.5 \\
EMNLP && & 1.2 & 2.6\\
COLING && & 0.6 & NA \\
Workshops && & -6.1 & -4.9 \\
misc && & -10.6 & -2.6  \\
1 early & 6.9 & 6.3 & 4.7 & 3.7 \\
2 early& 15.1 & 13.9  & 12.6 & 9.6  \\
3 early& 22.2 & 21.1 & 19.3 & 17.1   \\
4 early& 28.7 & 26.8& 26.3 & 20.8 \\
5 early& 33.3 & 32.4 & 27.8 & 27.3 \\
6 early& 38.3 & 36.9 & 33.5 & 31.7 \\
7 early& 42.1 & 40.9 & 38.1 & 34.3\\
8 early& 45.7 & 44.8 & 41.6 & 38.6 \\
9 early& 49.5 & 47.4 & 46.6 & 40.8 \\
10+ early & 59.4 & 59.8  & 54.6 & 54.1  \\
  \end{tabular}}
  \caption{Coefficients for regression (with $T=10$).  
  }
  \label{tab:coef}
\end{table}

This model performs a few simple transforms
on both the input and output variables:
\begin{enumerate}
    \setlength{\itemsep}{0pt}
    \setlength{\parskip}{0pt}
    \setlength{\parsep}{0pt}
    \item Percentile transform \cite{Bornmann2012TheUO,bornmann2014improve}:
    Predict percentiles instead of raw counts.
Percentiles are based on citations in fourth year after publication.
\item Thresholding transform: Since input citation counts have long tails,
we use \textit{pmin} to limit the number of factors in the regression to $T$.  
\end{enumerate}

Because the literature is growing exponentially \cite{bornmann2021growth}, care is required when comparing citations for papers published at different times \cite{Newman2013PredictionOH}.  We address these concerns by fitting coefficients
for each publication year.

Deep networks will likely produce better predictions, but
our goal here is to estimate the value of peer-review.
Is peer-review worth the cost, or should we publish more papers from ArXiv
with impressive early citations?



\autoref{tab:coef} shows regression coefficients
for $T=10$ and two publication dates (2016 and 2017).  The large set contains papers from PubMed, ArXiv and ACL.  The small set is for ACL venues.
The model produces coefficients for venues with 40 or more papers.  Venues with
less than 40 papers are assigned to \textit{misc}.  There is no coefficient
for COLING in 2017 because there was no COLING meeting in 2017.  To save space,
some venues were omitted from \autoref{tab:coef}.

As mentioned above, regression does not produce the best predictions in terms of loss,
but it has advantages in terms of interpretability.  
The coefficients on early citations in \autoref{tab:coef} show that more early
citations are better than fewer early citations.  Papers with 10+ early citations
are predicted to be in the $75^{th}$ percentile or better.  

The boxplots in \autoref{fig:ACL} 
show predictions from the model with $T=30$ for papers in the ACL Anthology published in 4 years
between 2016 and 2019.  
The coefficients are fit four times, once
for each publication year.  For each year,
predictions from the model are aggregated
by early citations (left) and by venue (right).

The width of the bars indicates the influence
of the other factor.  The bars are so narrow on the left that
they are hard to see, indicating that early citations are
very predictive of future citations.  Although venue
may be statistically significant, it has relatively little
consequence in practice.

These observations were confirmed by analysis of variance (ANOVA).
The ANOVA shows that early citations account for much more of the variance
than venue, as expected based on the discussion of correlations above.

Venues are sorted by median
predictions (computed over the papers published in that year).  While papers
published in more prestigious venues rank higher than papers in less prestigious venues,
the effect of venue is not only small, but also lacks robustness.
Note that the ordering
of venues varies from one year to the next:
EMNLP is in the top three venues in all four plots, though
it can be found in top place, second place and third place,
depending on the publication year.  Predictions based on early citations in \autoref{fig:ACL_citations} are more consistent over
the four publication years, indicating that early citations
are more robust than venue.  In particular, over all four panels,
there is a consistent trend for predictions to increase with 
the number of early citations.  The four panels in \autoref{fig:ACL_citations} are more similar to one another
than the four panels in \autoref{fig:ACL_venue}.


\section{Conclusions}
\subsection{Early Citations vs. Venue}
We showed that ``early returns'' (citations soon after publication)
are more predictive of future citations than venue.
This conclusion is based on:

\begin{enumerate}
    \setlength{\itemsep}{0pt}
    \setlength{\parskip}{0pt}
    \setlength{\parsep}{0pt}
    \item 
    \autoref{sec:cor}: Correlations ($\rho$)
    \item  \autoref{sec:h_and_mu}: h-index ($h$) and Impact ($\mu$)
    \item \autoref{sec:regression}: Regression
\end{enumerate}
\noindent

These observations suggest a simple actionable rule-of-thumb (use early returns)
that has advantages over current practice (reviewing) in terms
of exclusivity, inclusivity and robustness:

\begin{enumerate}
    \setlength{\itemsep}{0pt}
    \setlength{\parskip}{0pt}
    \setlength{\parsep}{0pt}
    \item
    \textbf{Exclusivity}:  Simple rule of thumb: for most venues, 
    1+ early citations are as good as reviews in terms of $\mu$;
   20+ early citations are better than reviews for most (all) venues.
    
\item \textbf{Inclusivity}: There are more papers ($N$) with 1+ early citations than in most (all) venues.
\item \textbf{Robustness}: Results were replicated over several sources of papers and publication dates.
\end{enumerate}



The rest of this paper will introduce
two controversial suggestions: (1) early citations
and (2) nominations to address two challenges (a)
too many submissions and (b) too few qualified reviewers.
Our goal is not so much to solve these challenges,
but merely to jump start a discussion
that might eventually lead to process improvements
that will scale better than the status quo.
We encourage the community, especially those
that do not like (1) and (2), to offer alternative
constructive suggestions.

\subsection{DDI Alternative to Reviewing}
\label{sec:DDI}


A number of challenges for reviewing were mentioned:
poorly defined tasks/incentives, validity, reliability,
subjectivity, biases, time, cost, scale and cheating.
Given these realities, is peer-reviewing worth the effort?
Are there faster, cheaper and more effective alternatives? 





\begin{enumerate}
    \setlength{\itemsep}{0pt}
    \setlength{\parskip}{0pt}
    \setlength{\parsep}{0pt}
    \item Open Peer-Review (OPR) \cite{List2017CrowdbasedPR}
    \item Don't Do It (DDI): Use early citations to reduce the load on peer-reviewing.
    \end{enumerate}
\noindent
Since OPR ``has neither a standardized definition nor an agreed schema of its features and implementations,'' \citet{RossHellauer2017WhatIO}, 
``{proposes a pragmatic definition of [OPR] as an umbrella term for... peer review models... including making reviewer and author identities open, publishing review reports and enabling greater participation...}.''

The DDI alternative is even more pragmatic and constructive.
Instead of reviewing papers, we suggest
the community post papers on ArXiv, and use early returns to help readers, authors and 
committees address questions such as:
\begin{enumerate}
    \setlength{\itemsep}{0pt}
    \setlength{\parskip}{0pt}
    \setlength{\parsep}{0pt}
    \item Readers: Who should read what?
    \item Authors: Who should cite what for what?
    \item Promotion and Award Committees:\\
    Who should be recognized for what?
\end{enumerate}

\subsection{New Role for Venues}

What should be the role for venues under this suggestion?
We suggest venues continue to publish high impact
papers in
their area that conform to their methods and practices,
but to do so in a way that copes more effectively with scale.
As mentioned above, the current system suffers from two concerns: (a) too many submissions and (b) too few qualified reviewers.
We suggest introducing a process upstream of program committees
to address both concerns.
To reduce the load, program committees should
focus on papers with impressive early citations, as well as
papers nominated by a process described below in section~\ref{sec:nomination}.



In addition to the first concern, reducing the load, these suggestions also help with the second concern, identifying
qualified/motivated reviewers.
It should be easier for those who have
cited the article to write a review since
they have already read the article and most
of the background material.  They are not
only better informed than a random reviewer,
but they are also probably more sympathetic to
the basic approach.

This proposal also simplifies the definition of the reviewing task.
By the time reviewers see the paper,
there is already considerable evidence of impact.
The question for reviewers becomes more about judging fit
than predicting impact.

\subsection{Nomination Process}
\label{sec:nomination}

In addition to early citations,
program committees should
accept nominations of papers to review from
thesis advisors and established researchers
in industrial research laboratories,
following precedents established
by nomination processes for awards such as ACM Doctoral Dissertation.\footnote{\url{https://awards.acm.org/doctoral-dissertation/nominations}}
To offset the reviewing load on society imposed by the nomination
process,
nominators should agree to review
four papers for each paper they nominate.
In this way, the proposed process
addresses both concerns raised above: (a) too many
submissions and (b) too few qualified/motivated reviewers.


\section{Ethics}

The proposed DDI method will not work with double-blind review, but people who have already cited the submission are unlikely to be biased against the submissions they have cited.

Mutual admiration societies have always existed in academia.  There is a danger that the proposed DDI method will encourage those practices.  However, citations leave an audit trail that makes it very easy for everyone to see what is happening.  As the cliche goes, sunlight is the best disinfectant.

Reviewing is a controversial topic.  From
the perspective of a conference organizer, we should
encourage controversial papers that engage the audience, and contribute significantly to the field.

\section{Limitations}

Citation counts can be gamed.  See discussion of cheating in \autoref{sec:cheating}.

This work is largely limited to English since the venues we consider emphasize English.

There is a risk that the proposed DDI/nomination method will help the rich get richer;
to compensate for this, there could be a process to encourage nominations from more diverse places.

\bibliography{custom}




\end{document}